\documentclass[10pt,twocolumn,letterpaper]{article}

\usepackage{iccv}
\usepackage{times}
\usepackage{epsfig}
\usepackage{graphicx}
\usepackage{amsmath}
\usepackage{amssymb}
\usepackage{multirow}
\usepackage{booktabs}

\usepackage{color,soul}
\usepackage{fancyhdr}
 
\fancypagestyle{plain}{
       \fancyhf{} 
       \fancyfoot[R]{\thepage} 

       \fancyfoot[L]{UNDER CONSIDERATION AT PATTERN RECOGNITION LETTERS, AUGUST 2019}

}

\usepackage[pagebackref=true,breaklinks=true,letterpaper=true,colorlinks,bookmarks=false]{hyperref}

\iccvfinalcopy 
\begin{document}

\title{BshapeNet: Object Detection and Instance Segmentation with Bounding Shape Masks}

\author{Ba Rom Kang\textsuperscript{2}, Ha Young Kim\textsuperscript{1,2,*}\\
\textsuperscript{1} Department of Financial Engineering, Ajou University\\
\textsuperscript{2} Department of Data Science, Ajou University\\
}

\maketitle

\begin{abstract}
Recent object detectors use four-coordinate bounding box (bbox) regression to predict object locations. Providing additional information indicating the object positions and coordinates will improve detection performance. Thus, we propose two types of masks: a bbox mask and a bounding shape (bshape) mask, to represent the object's bbox and boundary shape, respectively. For each of these types, we consider two variants: the “Thick” model and the “Scored” model, both of which have the same morphology but differ in ways to make their boundaries thicker. To evaluate the proposed masks, we design extended frameworks by adding a bshape mask (or a bbox mask) branch to a Faster R-CNN framework, and call this BshapeNet (or BboxNet). Further, we propose BshapeNet+, a network that combines a bshape mask branch with a Mask R-CNN to improve instance segmentation as well as detection. Among our proposed models, BshapeNet+ demonstrates the best performance in both tasks and achieves highly competitive results with state-of-the-art (SOTA) models. Particularly, it improves the detection results over Faster R-CNN+RoIAlign (37.3\% and 28.9\%) with a detection AP of 42.4\% and 32.3\% on MS COCO \texttt{test-dev} and Cityscapes \texttt{val}, respectively. Furthermore, for small objects, it achieves 24.9\% AP on COCO \verb'test-dev', a significant improvement over previous SOTA models. For instance segmentation, it is substantially superior to Mask R-CNN on both test datasets.
\end{abstract}
\section{Introduction}

An object detection algorithm determines the class and location of each object in an image. Deep learning-based approaches have  achieved notable success recently, such as the Overfeat \cite{sermanet2013overfeat}, Fast R-CNN \cite{girshick2015fast}, and Faster R-CNN \cite{ren2015faster}. These methods use a bounding box (bbox) regressor to predict the object locations that are defined by four-dimensional coordinates (x, y, width, and height). However, it is not easy to learn continuous variables from images. Thus, if we define a new target that allows the detector to learn the position of the object more efficiently and add it to the existing framework, the performance will improve. In other words, the algorithm can predict the position more accurately by learning not only the coordinates but also a different form of location information. This is because of the same reason why people learn better if they study the same things differently.

In this study, we define the location of an object in the form of a mask. This is because we perceive that spatial information can be learned more efficiently than coordinates, and can facilitate not only object detection but also instance segmentation such as Mask R-CNN \cite{he2017mask}. Because the object’s boundary separates the foreground and background, we consider it to be more crucial than the object’s interior. Thus, we  transformed the complex task of learning both the interior and boundary into a simpler task by focusing only on the boundaries. In particular, we consider it highly effective for small objects and occlusions.

Thus, we propose two types of masks: a \textit{bbox mask} and a \textit{ bounding shape (bshape) mask}, to indicate the location of an object. Figures 1 (a) and (e) show masks with only true boundaries; however, an imbalance problem occurs owing to excessive zeros. Therefore, it is necessary to create a thick boundary. For each of the two types, we consider two variants: the “\textit{Thick}” model (Figures 1 (b) and (f)) and the “\textit{Scored}” model (Figures 1 (c) and (g)), both of which exhibit the same morphology but differ in ways to make their boundaries thicker. Furthermore, as shown in Figures 1(d) and (h), various thickness masks are used.  

\begin{figure*}
  \centering    
  \includegraphics[width=160mm,height=56mm]{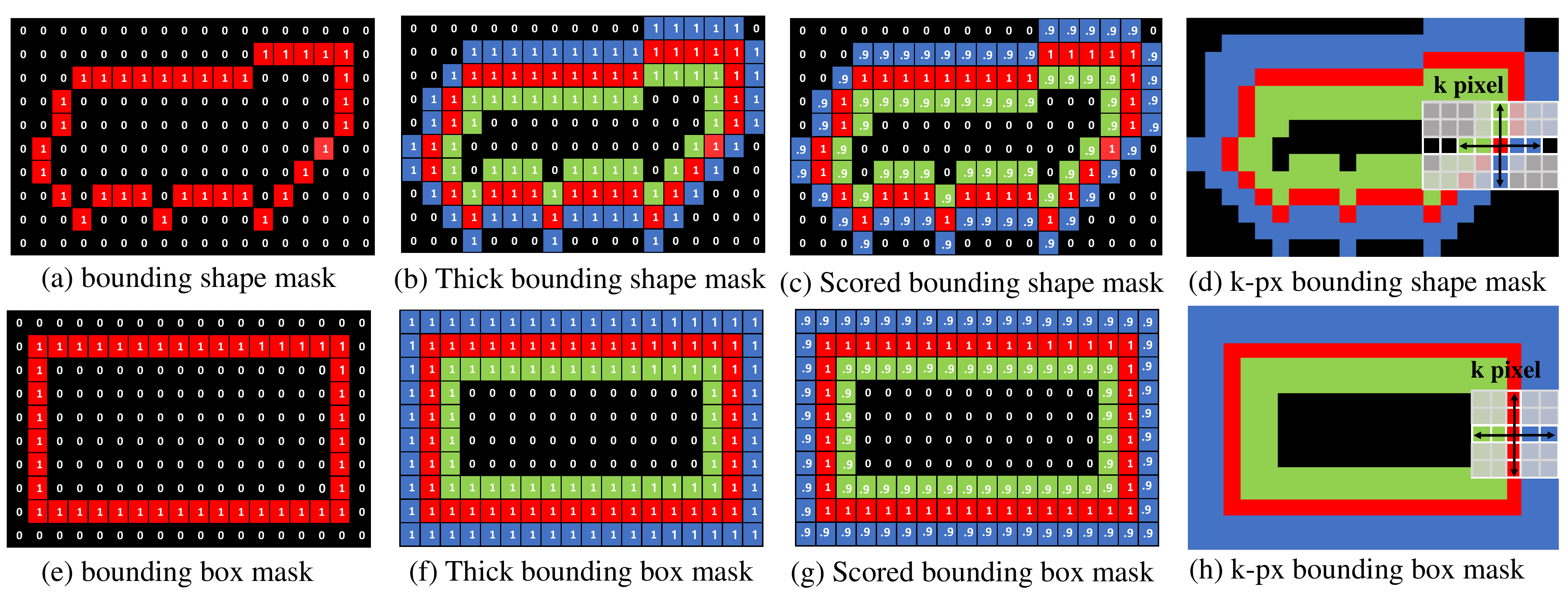}
  \caption{Proposed bounding shape masks and bounding box masks.}
\end{figure*}
We propose three frameworks, \textit{BboxNet}, \textit{BshapeNet}, and \textit{BshapeNet+}, to verify the newly defined masks (Figure 2) for object detection and instance segmentation. BshapeNet (or BboxNet) is an extended framework by adding a bshape mask (or a bbox mask) branch to the Faster R-CNN framework with RoIAlign \cite{he2017mask}, called Faster R-CNN$\ddagger$. BshapeNet+ is a network in which a bshape mask branch and instance mask branch (adopted from Mask R-CNN \cite{he2017mask}) are added to Faster R-CNN$\ddagger$. It is the same as combining a bshape mask branch with the Mask R-CNN. With this network,  the effect of the bshape mask can be analyzed in terms of instance segmentation and object detection when using the bshape mask and instance mask together. To clarify again, BshapeNet(+) allows both object detection and instance segmentation, but BboxNet only allows object detection because the mask predicts the bounding boxes. Further, each network contains two variants (the Thick model and the Scored model).

We evaluate our approaches on MS COCO \cite{lin2014microsoft} and the Cityscapes \cite{cordts2016cityscapes} datasets. The main contributions of this study are summarized as follows:\\
(1) We propose novel and efficient frameworks, BshapeNet and BshapeNet+, by adding a newly introduced bshape mask branch to the Faster R-CNN$\ddagger$ and Mask R-CNN, for improving object detection and instance segmentation. Our masks are extremely easy to add to the existing algorithms and can be removed at inference. \\
(2) We propose and investigate two types of masks, and their various variants (Scored mask, Thick mask, and the degree of boundary dilation). Thus, we demonstrate that Scored BshapeNet+ exhibits the best performance and that our methods are also effective for small objects. Further, we confirm the possibility of replacing the bbox regressor by the bbox mask through BboxNet.\\
(3) Experiments on two benchmarks demonstrate that our Scored BshapeNet+ improves performance significantly compared to the Faster R-CNN$\ddagger$ and Mask R-CNN, and achieves highly competitive results with SOTA methods.

\section{Related Work}
\noindent\textbf{Object Detection:}\hspace{.5em}Many studies have recently presented excellent results in object detection \cite{liu2016ssd,ren2015faster,redmon2016you}. The remarkable improvements in object detection began with Overfeat and the region-based CNN (R-CNN) \cite{girshick2014rich}. In particular, the family of R-CNNs proposed in 2014 remains outstanding. Unlike sliding-window methods such as Overfeat, the R-CNN selects proposals including objects, and subsequently localizes objects using the selected proposals. To solve the heavy calculations of the R-CNN, the Fast R-CNN and Faster R-CNN were developed. The Fast R-CNN uses a \textit{selective search} \cite{uijlings2013selective} to obtain proposals using the features of the CNN backbone, and a region of interest (RoI) pooling layer \cite{girshick2015fast} to eliminate the repeated feeding of RoIs back into the CNN. The Faster R-CNN uses region proposal networks (RPNs) to detect RoIs in the structure of a Fast R-CNN. YOLO \cite{redmon2016you} and SSD \cite{liu2016ssd} have no additional networks to detect the RoIs, and perform object proposal and object detection simultaneously to reduce calculations.

 \begin{figure*}
  \centering    
  \includegraphics[width=135mm]{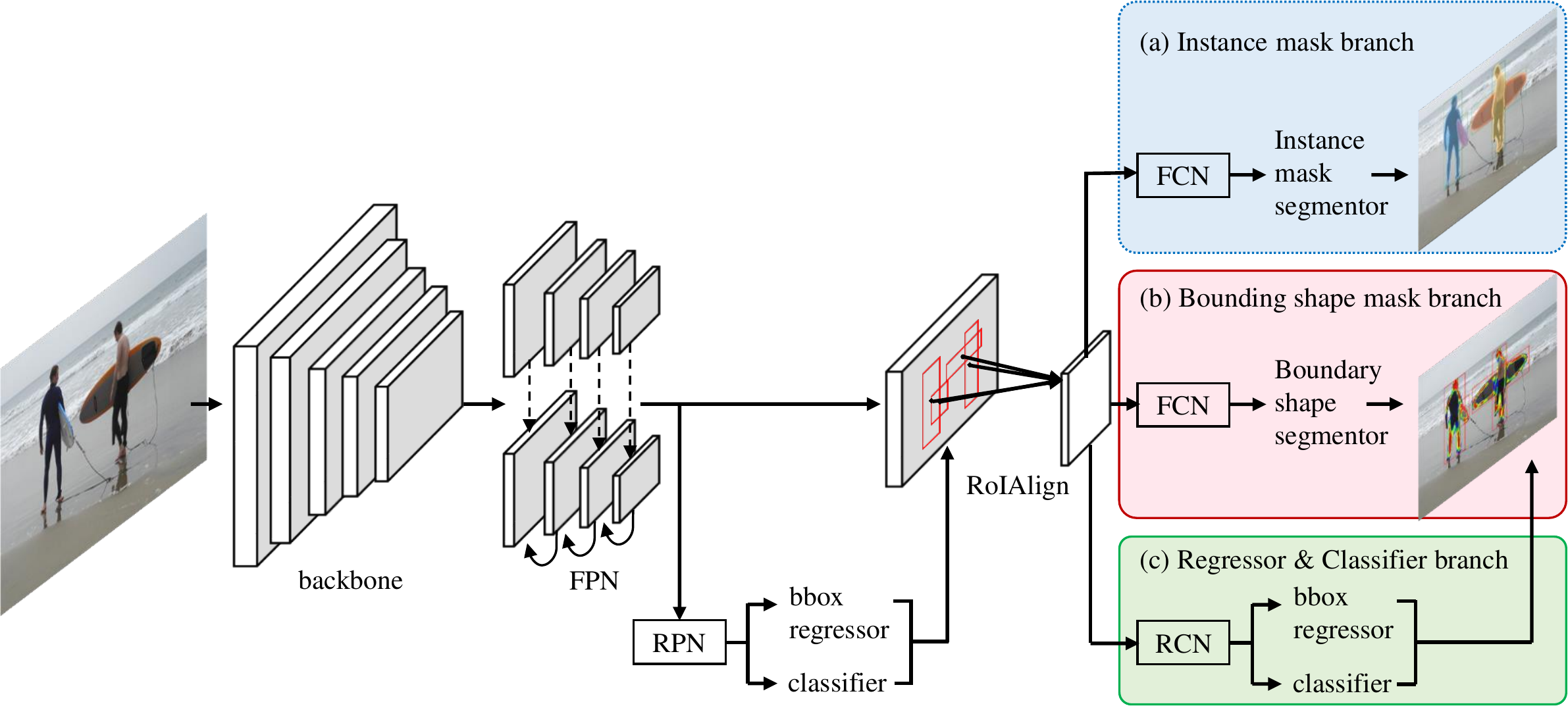}
  \caption{The Proposed BshapeNet+ framework for object detection and instance segmentation.}
\end{figure*}

\noindent\textbf{Instance Segmentation:}\hspace{.5em} 
DeepMask \cite{pinheiro2015learning} predicts the location of an object by segmenting an object to propose candidates. Dai \textit{et al.} \cite{dai2016instance} proposed a cascading stage model using a shared feature to segment the proposed instance(s) from a bbox. The key idea of a fully convolutional instance-aware semantic segmentation (FCIS) \cite{li2016fully} is to predict a position-sensitive output through a fully convolutional structure. This is a combination of the segment proposal method of a previous study \cite{dai2016instance2} and the object detection method of \cite{dai2016instance}. FCIS predicts the bbox and mask simultaneously, but exhibits a systemic error with overlapping objects and only detects the bounds of objects approximately. The Mask R-CNN uses an elaborate instance segmentation result. This model uses RoIAlign \cite{he2017mask} to obtain precise RoIs and uses them together with three branches (classification, bbox detection, and segmentation) to achieve a satisfactory performance. These methods focus on the inside of the proposed object for instance segmentation. In contrast, we study the edges of objects (information that distinguishes between the foreground and background).
\section{Our Approach}
\noindent\textbf{Bshape (or Bbox) Mask Representation:}\hspace{.5em}We define two types of masks, bshape mask and bbox mask, for better object detection and instance segmentation, as in Figure 1. We defined these masks for two primary reasons. The first reason is as follows. When people learn, they learn better by listening and observing rather than only by listening. Similarly, if an object detector learns not only the coordinates but also a different form of position of the object with our proposed mask, then it localizes the objects better. Another reason is that if we create a method to improve or support the instance mask of the Mask R-CNN, a leading instance segmentation method, performance will be improved.

As the names of the proposed masks imply, the bshape mask labels the boundary pixels of the object as 1, while the bbox mask labels the bbox pixels as 1 and the other pixels as zero (see Figures 1(a) and (e)). With these masks, it is difficult to learn because of excessive zeros compared to ones. Thus, for each of the bshape mask and bbox mask, we define two variants, the Thick mask and the Scored mask. They exhibit the same morphology but differ in methods to thicken their boundaries (or bboxes). We create thicker boundary by extending the boundaries inside and outside of the true boundary by $k$ pixels. We prefix the name of these masks with "\textit{$k$-px}" to indicate the boundary thickness. We call the expanded boundary pixels \textit{false boundary pixels}. We explain the Thick mask and Score mask with the bshape mask in more detail below. The bbox mask is the same except that the bounding box pixels are considered.

In the Thick mask, false boundary pixels are filled with 1s, as is the case for the true boundary pixels. The mathematical expression of this mask is as follows. Let \(M\) \(=\) \((m_{ij})\) \(\in\) \(\mathbb{R}^{v \times w}\) be a true boundary mask matrix, \(B\) \(=\lbrace (i,j):\) \(m_{ij} = 1,\) \(1\leq i \leq v, 1\leq j\leq w\rbrace\) is a true boundary index set, and \(X\) \(=\) \((x_{pq})\) \(\in\) \(\mathbb{R}^{v \times w}\) is a \(k\)-px Thick bshape mask matrix. For $\forall$\((i,j)\) \(\in\) \(B\),
\begin{eqnarray}
\left\{\begin{array}{rcl}
                     x_{pj} = 1, &if &1\leq i - k \leq p \leq i + k \leq v\\
                     x_{iq} = 1, &if &1\leq j - k \leq q \leq j + k \leq w.\\
\end{array}\right.
\end{eqnarray}
All remaining values are filled with zeros. In this case, it is difficult to distinguish between the false and true boundaries; therefore, we develop the Scored mask because we can learn the boundaries more effectively by defining false boundary and true boundary values differently. The value of the false boundary is reduced at a constant rate in proportion to the distance from the actual boundary. Thus, for the Scored bounding shape mask, false boundary pixels are filled with distance-based scored numbers, which are positive numbers less than 1, to generate a mask. Let \(Y\) \(=\) \((y_{pq})\) \(\in\) \(\mathbb{R}^{v \times w}\) be a \(k\)-px Scored bounding shape (or box) mask matrix. We use the same matrices \(M\) and \(B\) as in \(Eq. 1\), and \(s\) is a predetermined positive constant (less than 1) that controls the magnitude by which the value decreases. We set $s$ as 0.05. Subsequently, $\forall$\((i,j)\) \(\in\) \(B\),
\begin{eqnarray}
\left\{\begin{array}{rcl}
                     y_{pj} = 1-d_{1}s, &if &1\leq i - k \leq p \leq i + k \leq v\\
                     y_{iq} =  1-d_{2}s, &if &1\leq j - k \leq q \leq j + k \leq w,\\
\end{array}\right.
\end{eqnarray}
where \(d_{1}=\left|p-i\right|\) and \(d_{2}=\left|q-j\right|\) are the distances from the true boundary, and the remaining pixels are zero.


\noindent\textbf{Proposed Frameworks:}\hspace{.5em}To verify our masks, we developed three frameworks: BboxNet, BshapeNet, and BshapeNet+, in this study (Figure 2). The BshapeNet (or the BboxNet) is a framework that combines the Faster R-CNN$\ddagger$ with a bshape (or bbox) mask branch. In other words, we replaced the instance mask branch in the Mask R-CNN with the bshape (or bbox) mask branch. Further, BshapeNet+ is a framework that adds both our bshape mask branch and instance mask branch to the Faster R-CNN$\ddagger$. In other words, it is the same as combining a bshape mask with the Mask R-CNN. The bshape (or bbox) mask branch segments the boundaries (bboxes) of instances in the RoIs. Further, the instance mask branch performs instance segmention, that is, the interior of the instance is segmented as well as the boundary \cite{he2017mask}. The regressor and classifier branch perform bbox regression and classification of the RoIs used in \cite{girshick2015fast, ren2015faster, lin2017feature}. For clarity, Table 1 summarizes the branches used for each model.

As shown in Figure 2, BshapeNet+ consists primarily of a backbone (as in \cite{ren2015faster, he2017mask}), the RPN, the region classification network (RCN), and the bshape mask branch and instance mask branch based on the FCN \cite{long2015fully}. In addition, for better performance, we used the feature pyramid network (FPN) \cite{lin2017feature}. More specifically, the flow of BshapeNet+ with each component is as follows. First, the backbone  extracts features and then the FPN combines multiresolution features with high-level features from multiple layers of the backbone \cite{lin2017feature} and forwards combined features to the RPN. Subsequently, the classifier and bbox regressor of the RPN propose the RoIs. For final predictions, both the bshape segmentor and instance segmentor use RoIs simultaneously as the RCN. Through this process, all predictions occur.
\begin{table}[t]
	\begin{center}
		\resizebox{\linewidth}{!}{ 
			\begin{tabular}{l|l|c|c}
				\hline\noalign{\smallskip}
				Model&\multicolumn{1}{c|}{Training} &Test (Detection) &Test (Segmentation)  \\
				\noalign{\smallskip}
				\hline
				\noalign{\smallskip}
				BboxNet&(b)$\dagger$ bbox, (c) regressor&(c) regressor&-\\
				\noalign{\smallskip}
				BshapeNet&(b) bshape, (c) regressor&(c) regressor&(b) bshape \;\\
				\noalign{\smallskip}
				BshapeNet+&(a) instance, (b) bshape, (c) regressor&(c) regressor&(a) instance\\				
				\noalign{\smallskip}
				\hline
		\end{tabular}}	
	\end{center}
	\caption{The branches used in each proposed model in Figure 2 for training and test. b$\dagger$ denotes the bbox mask branch.}
\end{table}

Figure 3 shows the architecture of the BshapeNet. It is exactly the same as the architectures of the Mask R-CNN and BboxNet. BshapeNet+ contains two FCNs with the same architecture. In more detail, we investigated our models with ResNet \cite{he2016deep} and ResNeXt \cite{xie2017aggregated} as the backbone for all experiments and these are composed in six stages, where each stage consists of CNN layers of equal feature map size. The feature maps finally extracted at each stage are called C1, C2, C3, C4, C5, and C6. Among them, C2, C3, C4, and C5 were fed to the FPN and P2, P3, P4, and P5 were output, respectively. The architectures of the FCN used in the bshape mask branch and instance mask branch are the same as that of the Mask R-CNN. 

\begin{figure}[!]
  \centering    
  \includegraphics[width=85mm,height=32mm]{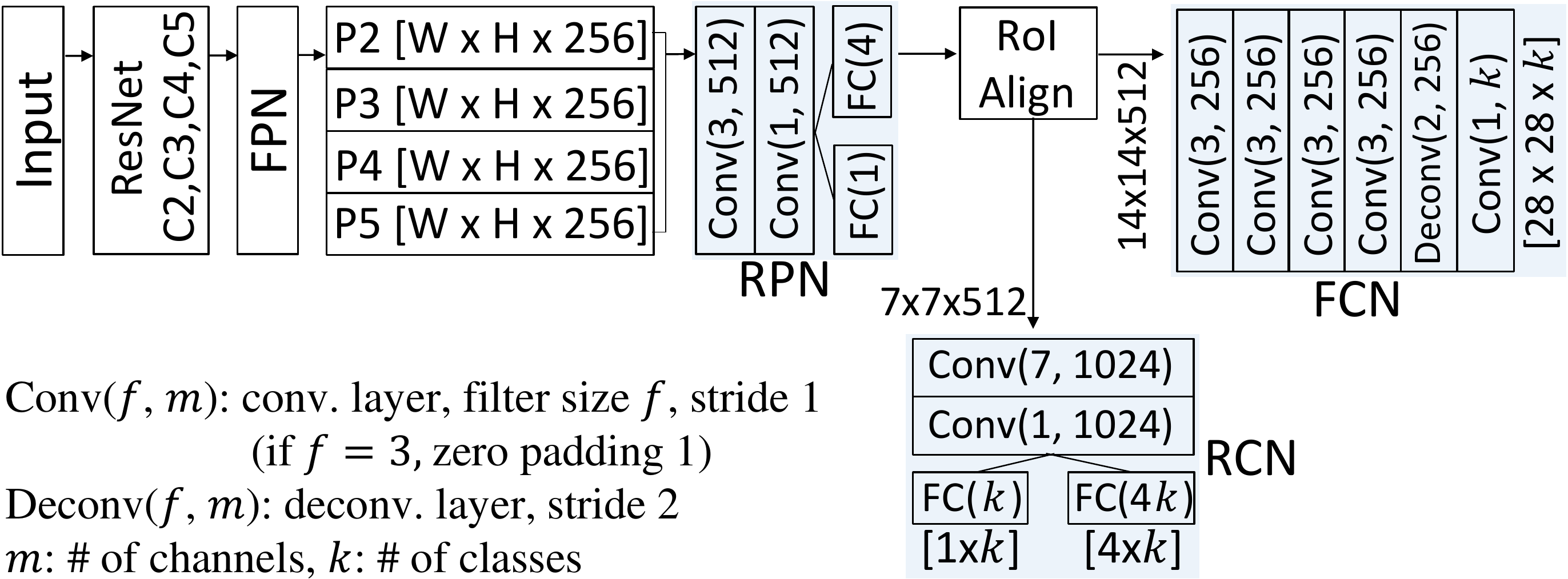}
  \caption{The detailed network architecture of BshapeNet.}
\end{figure}

\noindent\textbf{Training Loss Functions:}\hspace{.5em}We used the same loss for both BshapeNet and BboxNet because only the morphology of the defined mask was different. However, the Scored model and Thick model exhibit different losses. First, the loss function for the Scored model is defined as follows:
\begin{equation}
  Loss_{total}= \alpha L_{RPN}+\beta L_{RCN}+\gamma L_{Smask},
\end{equation}
where \(\alpha , \beta\) and \(\gamma\) are predetermined positive constants, and the loss functions of the RPN, RCN, and Scored bshape (or bbox) mask branch are called \(L_{RPN}\), \(L_{RCN}\), \(L_{Smask}\), respectively. The loss functions of the RPN and RCN are the same as those for the faster R-CNN. 

We used the Euclidean loss function for the Scored mask branch, called $L_{Smask}$, as follows: for each RoI, $L_{Smask}=\frac{1}{2HW}\sum_{i}^{W}\sum_{j}^{H}(t_{ij}-\hat{t}_{ij})^2$, 
where \(H\) and \(W\) indicate the height and width of the mask, respectively, and \(t_{ij}\) is the ground-truth label of $(i, j)$ in the mask with a predicted value of \(\hat{t}_{ij}\).
The Thick mask branch solves a pixel-wise classification while the Scored mask branch model solves a pixel-wise regression. 

Thus, we used the following binary cross-entropy loss function for the Thick mask branch, called $L_{Tmask}$, with the same notation as in  $L_{Smask}$: $L_{Tmask}=-\frac{1}{HW}\sum_{i}^{W}\sum_{j}^{H}\lbrace t_{ij}\log \hat{t}_{ij}+(1-t_{ij})\log(1-\hat{t}_{ij})\rbrace$. The total loss function of the Thick model is equivalent to changing \(L_{Smask}\) to \(L_{Tmask}\) in Eq. 3. 

Finally, the loss function of BshapeNet+ is equivalent to adding \(\delta L_{Imask}\) to Eq. 3. $L_{Imask}$ stands for the loss of the instance mask branch and is the same as that used in the Mask R-CNN and $\delta$ is a preset positive constant.

\noindent\textbf{Inference:}\hspace{.5em}Except for Table 8, to analyze the effect of adding our mask to the existing model, at inference, we evaluate the performance using the branches, as shown in Table 1. The performance of bbox mask branch is also evaluated (Table 8). Simple post-processing is required for instance segmentation with our bshape mask branch because our bshape masks only segment the boundaries of objects. We performed post-processing for the two-step procedures. The first step is to connect the predicted boundary and the second step is to fill it. To connect the predicted bshape mask, we used a modified Prim’s algorithm \cite{prim1957shortest}, which is a technique used in the minimum spanning tree \cite{gower1969minimum}. 

\section{Experiment}
We compare the results of our models, BshapeNet+, BshapeNet, and BboxNet with their variants, and perform ablations. We also compare these results with the Faster R-CNN$\ddagger$, Mask R-CNN, and SOTA algorithms. We used two benchmark datasets, MS COCO and Cityscapes. Specifically, we used the MS COCO Detection 2017 dataset containing 81 classes and 123,287 images  (\verb'trainval'), and 118,287 images for the training set and 5K images for the validation set (\verb'minival'). We also obtained the results of object detection and instance segmentation on \verb'test-dev' \cite{lin2014microsoft}. For Cityscapes, we used the dataset with \verb'fine' annotations comprising nine object categories for instance-level semantic labeling and a 2,975-image training set, 500-image validation set, and 1,525-image test set \cite{cordts2016cityscapes}. Further, we evaluated our results using the Cityscapes \verb'test-server'. We evaluated the performance using the standard COCO-style metric \cite{lin2014microsoft}. 

In the Mask R-CNN paper, models were trained using 8 NVIDIA Tesla P100 GPUs; however, we used 2 NVIDIA GTX 1080Ti GPUs. Thus, we used a smaller minibatch. Owing to the limited experimental environment, we re-experiment the Mask R-CNN and Faster R-CNN$\ddagger$ models in our experimental environment for a fair comparison. Further, we matched the hyperparameters to the paper of Mask R-CNN except the minibatch size.  
\begin{table}[t]
	\begin{center}
		
		\resizebox{\linewidth}{!}{ 
			\begin{tabular}{c|l|lll|l|lll}

				\hline\noalign{\smallskip}
				\multicolumn{2}{l|}{Model (R-101)}&  \(AP_{bb}\) &\(AP^{50}_{bb}\) &\(AP^{75}_{bb}\) & Model(R-101) & \(AP_{bb}\) &\(AP^{50}_{bb}\) &\(AP^{75}_{bb}\) \\
				\noalign{\smallskip}
				\hline
				\noalign{\smallskip}
				\multirow{5}{*}{{\rotatebox[origin=c]{90}{\large{T h i c k}}}}
				&BboxNet (3)  &\Large37.9&\Large59.9&\Large40.6&BshapeNet (3)  &\Large38.1&\Large59.9&\Large41.0\\
				\noalign{\smallskip}
				&BboxNet (5)  &\Large37.9&\Large59.0&\Large40.0&BshapeNet (5) &\Large38.2&\Large60.9&\Large40.5\\
				\noalign{\smallskip}
				&BboxNet (7) &\Large38.0&\Large59.9&\Large39.4&BshapeNet (7) &\Large38.4&\Large61.5&\Large41.8\\
				\noalign{\smallskip}
				&BboxNet (11)  &\Large37.8&\Large58.0&\Large39.8&BshapeNet (11)&\Large38.2&\Large60.9&\Large41.2\\
				\noalign{\smallskip}
				\hline
				\noalign{\smallskip}
				\multirow{7}{*}{{\rotatebox[origin=c]{90}{\Large{S c o r e d}}}}
				&BboxNet (3) &\Large37.8&\Large59.5&\Large40.5&BshapeNet (3) &\Large41.5&\Large63.4&\Large44.0\\
				\noalign{\smallskip}
				&BboxNet (5) &\Large38.1&\Large59.8&\Large40.7&BshapeNet (5) &\Large41.4&\Large63.5&\Large45.9\\
				\noalign{\smallskip}
				&BboxNet (7) &\Large38.1&\Large59.7&\Large40.9&\textbf{BshapeNet (7)}&\Large\textbf{42.1}&\Large\textbf{64.1}&\Large\textbf{46.2}\\
				\noalign{\smallskip}
				&BboxNet (11) &\Large37.9&\Large59.5&\Large40.5& BshapeNet (11)&\Large41.7&\Large63.9&\Large46.2\\
				\noalign{\smallskip}
				\cline{2-9}
				\noalign{\smallskip}
				 &BshapeNet+ (3) &\Large41.7&\Large63.3&\Large44.6&\textbf{BshapeNet+ (7)}  &\Large\textbf{42.3}&\Large\textbf{64.5}&\Large\textbf{46.4}\\
				 \noalign{\smallskip}
				&BshapeNet+ (5)  &\Large41.8&\Large63.8&\Large46.2 &BshapeNet+ (11)  &\Large42.1&\Large64.1&\Large46.3\\
				 \noalign{\smallskip}
				\hline
				\noalign{\smallskip}
				\multicolumn{2}{c|}{Faster R-CNN$\ddagger$ \small{(ours)}}&\Large37.0&\Large58.8&\Large36.3&Mask R-CNN \small{(ours)}&\Large38.0&\Large60.1&\Large41.5\\
				
				\noalign{\smallskip}
				\hline
		\end{tabular}}	
		
	\end{center}
	\caption{Object detection results on COCO \texttt{minival} dataset with bbox AP (\%). Numbers in parentheses indicate thickness of boundary ($k$-px model). We denote ResNet by ``R'' and \textit{ours} means results obtained from our experimental environment.}
		
\end{table}
\begin{table}
\begin{center}

\resizebox{\linewidth}{!}{
\begin{tabular}{c|l|ccc|l|ccc} 
\hline\noalign{\smallskip}
\multicolumn{2}{l|}{Model (R-101)}& \(AP_{mk}\) &\(AP^{50}_{mk}\) &\(AP^{75}_{mk}\) &Model (R-101)& \(AP_{mk}\) &\(AP^{50}_{mk}\) &\(AP^{75}_{mk}\) \\
\noalign{\smallskip}
\hline
\noalign{\smallskip}\noalign{\smallskip}
\multirow{2}{*}{{\rotatebox[origin=c]{90}{T h i c k}}}
&BshapeNet (3)  &\Large{30.7}&\Large{48.6}&\Large{27.9}&BshapeNet (7) &\Large{31.5}&\Large{51.6}&\Large{31.9}\\
\noalign{\smallskip}\noalign{\smallskip}
&BshapeNet (5) &\Large{31.2}&\Large{52.2}&\Large{31.4}&BshapeNet (11)&\Large{31.6}&\Large{50.4}&\Large{29.8}\\		                                                                            
\noalign{\smallskip}\noalign{\smallskip}
\hline
\noalign{\smallskip}
\multirow{6}{*}{{\rotatebox[origin=c]{90}{ S c o r e d}}}
&BshapeNet (3) &\Large{33.2}&\Large{49.8}&\Large{30.3}&\textbf{BshapeNet (7)}&\Large{\textbf{36.7}}&\Large{\textbf{57.7}}&\Large{\textbf{38.0}}\\
\noalign{\smallskip}\noalign{\smallskip}
&BshapeNet (5) &\Large{34.8}&\Large{57.1}&\Large{37.6}&BshapeNet (11)&\Large{36.4}&\Large{57.2}&\Large{37.9}\\
\noalign{\smallskip}
\cline{2-9}
\noalign{\smallskip}
&BshapeNet+ (3)  &\Large{33.6}&\Large{50.1}&\Large{32.6}&\textbf{BshapeNet+ (7)}  &\Large{\textbf{37.1}}&\Large{\textbf{58.9}}&\Large{\textbf{39.3}}\\
\noalign{\smallskip}\noalign{\smallskip}
& BshapeNet+ (5)  &\Large{35.4}&\Large{57.4}&\Large{37.9}&BshapeNet+ (11)  &\Large{36.9}&\Large{57.7}&\Large{38.9}\\
 \noalign{\smallskip}
\hline
\noalign{\smallskip}
\multicolumn{2}{l|}{Mask R-CNN \cite{he2017mask}}&\Large{35.4}&\Large{57.3}&\Large{37.5}&Mask R-CNN \small{(ours)}&\Large{35.2}&\Large{57.3}&\Large{37.6}\\
\noalign{\smallskip}
\hline
\end{tabular}}

\end{center}
\caption{Instance segmentation results on COCO \texttt{minival} dataset with bbox AP and mask AP (\%). }
\end{table}

\begin{figure*}[!t]
  \centering    
  \includegraphics[width=150mm,height=34mm]{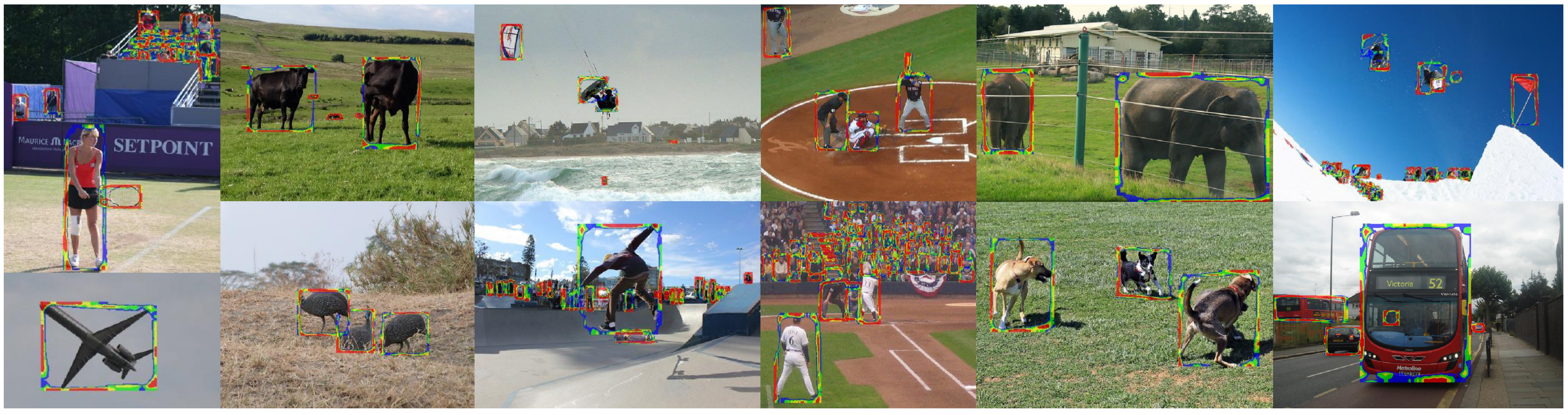}
  \caption{Examples of bounding box segmentation results of 7-px Scored BboxNet (ResNet101) on the MS COCO \texttt{minival} dataset. No post-processing has been performed.}
\end{figure*}

\begin{table*}[!t]
	\begin{center}
		
		\resizebox{15.5cm}{!}{
			\begin{tabular}{l|c|c|cccccc|cccccc}
				\hline\noalign{\smallskip}
				Model&test &backbone&\(AP_{bb}\)&\(AP^{50}_{bb}\) &\(AP^{75}_{bb}\) &\(AP^{L}_{bb}\) &\(AP^{M}_{bb}\)&\(AP^{S}_{bb}\) &\(AP_{mk}\)&\(AP^{50}_{mk}\) &\(AP^{75}_{mk}\) &\(AP^{L}_{mk}\) &\(AP^{M}_{mk}\) &\(AP^{S}_{mk}\) \\
				\noalign{\smallskip}
				\hline
				\noalign{\smallskip}
				Mask R-CNN (ours)&minival  &R-50&\large37.8&\large58.0&\large38.3&\large49.2&\large40.1&\large19.1&\large33.4&\large54.7&\large33.9&\large50.1&\large37.0&\large14.9\\
				\noalign{\smallskip}
				+ deeper (ours)&minival  &R-101&\large38.0&\large60.1&\large41.5&\large50.1&\large40.7&\large19.7&\large35.2&\large57.3&\large37.6&\large51.3&\large37.1&\large15.5\\
				\noalign{\smallskip}
			     + ResNeXt (ours)&minival &X-101&\large39.3&\large61.1&\large43.0&\large51.0&\large44.0&\large23.6&\large36.7&\large58.5&\large38.7&\large52.1&\large38.5&\large16.5\\
				\noalign{\smallskip}				
				\hline
				\noalign{\smallskip}	
				Faster R-CNN$\ddagger$ (ours)&minival &R-50&\large35.9&\large57.4&\large35.1&\large48.0&\large39.2&\large17.3&\large-&\large-&\large-&\large-&\large-&\large-\\
				\noalign{\smallskip}
				BshapeNet  &minival &R-50&\large38.2&\large60.6&\large40.8&\large50.5&\large41.8&\large22.6&\large35.8&\large54.4&\large36.3&\large50.9&\large37.3&\large16.2\\
				\noalign{\smallskip}
				+ deeper &minival  &R-101&\large42.1&\large64.1&\large46.2&\large52.8&\large44.7&\large24.9&\large36.7&\large57.7&\large38.0&\large51.3&\large37.7&\large16.5\\
				\noalign{\smallskip}
			     + ResNeXt&minival &X-101&\large42.3&\large64.4&\large46.7&\large52.9&\large45.1&\large25.1&\large37.0&\large58.9&\large38.7&\large52.5&\large38.6&\large16.8\\
			     \noalign{\smallskip}
			    + Inst. mask (BshapeNet+)&minival &X-101&\large\textbf{42.5}&\large\textbf{64.7}&\large\textbf{46.9}&\large\textbf{53.1}&\large\textbf{45.6}&\large\textbf{25.2}&\large\textbf{37.5}&\large\textbf{59.5}&\large\textbf{39.7}&\large\textbf{54.1}&\large\textbf{40.3}&\large\textbf{17.4}\\
			    \noalign{\smallskip}	\toprule
				Mask R-CNN (ours)&test-dev  &R-101&\large38.0&\large60.2&\large41.4&\large50.0&\large41.4&\large20.1&\large35.5&\large57.4&\large37.9&\large52.1&\large37.8&\large15.1\\
				\noalign{\smallskip}
			     + ResNeXt (ours)&test-dev &X-101&\large39.5&\large61.3&\large43.1&\large51.2&\large43.2&\large22.1&\large36.9&\large58.5&\large38.9&\large52.7&\large38.6&\large16.9\\
				\noalign{\smallskip}
				Mask R-CNN \cite{he2017mask}&test-dev   &R-101&\large38.2&\large60.3&\large41.7&\large50.2&\large41.1&\large20.1&\large35.7&\large58.0&\large37.8&\large52.4&\large38.1&\large15.5\\
				\noalign{\smallskip}
			     + ResNeXt \cite{he2017mask}&test-dev &X-101&\large39.8&\large62.3&\large43.4&\large51.2&\large43.2&\large22.1&\large37.1&\large60.0&\large39.4&\large53.5&\large39.9&\large16.9\\
				\hline
				\noalign{\smallskip}
				Faster R-CNN$\ddagger$ \cite{he2017mask} &test-dev &R-101&\large37.3&\large59.6&\large40.3&\large48.8&\large40.2&\large19.8&\large-&\large-&\large-&\large-&\large-&\large-\\
				\noalign{\smallskip}				
				BshapeNet  &test-dev &R-101&\large42.3&\large64.5&\large46.4&\large53.0&\large45.7&\large24.9&\large37.0&\large58.1&\large38.2&\large51.9&\large38.5&\large16.7\\
				\noalign{\smallskip}
			     + ResNeXt&test-dev &X-101&\large42.5&\large64.8&\large46.7&\large53.1&\large45.7&\large25.1&\large37.2&\large58.9&\large38.9&\large52.7&\large38.9&\large16.9\\
			     \noalign{\smallskip}
			    + Inst. mask (BshapeNet+)&test-dev &X-101&\large\textbf{42.8}&\large\textbf{64.9}&\large\textbf{46.9}&\large\textbf{53.6}&\large\textbf{46.1}&\large\textbf{25.2}&\large\textbf{37.9}&\large\textbf{61.3}&\large\textbf{40.2}&\large\textbf{54.4}&\large\textbf{40.4}&\large\textbf{17.4}\\
				\noalign{\smallskip}
				\hline
				\noalign{\smallskip}
		\end{tabular}}
	\end{center}
	\caption{Ablation study on BshapeNet on COCO \texttt{minival} and \texttt{test-dev}. We denote ResNeXt by ``X'' for brevity. The ``Inst. mask'' means that an instance mask branch; The 7-px Scored BshapeNet is used.  }
\end{table*}

\noindent\textbf{Metric:}\hspace{.5em}We follow the standard MS COCO evaluation metrics including \(AP\) (average precision averaged over intersection-over-union (IoU) thresholds of 0.5 to 0.95), \(AP^{50}\) (IoU = 0.5), \(AP^{75}\) (IoU = 0.7), and \(AP^S\), \(AP^M\), \(AP^L\), which are APs for small, medium, and large objects, respectively. We specified the bbox average precision as \(AP_{bb}\) and instance segmentation average precision as \(AP_{mk}\). These metrics apply to both datasets.

\noindent\textbf{Implementation details:}\hspace{.5em} The detailed architectures of our models are described in the previous section. For the MS COCO dataset, we resized the images such that their shorter edge was 800 pixels \cite{lin2017feature}. We used two GPU and four minibatch (two images per GPU) and trained the model for 64K iterations.  We set the initial learning rate to 0.02 and divided it by 10 (0.002) at 48k iterations. For the backbone, we used ResNet50, ResNet101, and ResNeXt101. 

In the Cityscapes, we performed training using only a \verb'fine' dataset. Although the raw data size was 2048\(\times\)1024, we reduced it to 1024\(\times\)800 to fit our resource. The models were trained with two GPUs and two minibatches (one image per GPU), and the model for 96K iterations was trained. We set the initial learning rates to 0.01 divided by 10 (0.001) at 56K iterations and used only ResNet50 as the backbone. This is because the amount of data in Cityscapes are extremely small and our model does not improve significantly with ResNet101 as the Mask R-CNN \cite{he2017mask}.

The hyperparameters typically used in both datasets are as follows. Each image contains 512 sampled RoIs for the FPN with positive and negative 1:3 ratios \cite{lin2017feature}. Anchors of five different scales and three aspect ratios were used as in \cite{ren2015faster}. The proposed RoIs of the RPN are 2000 per image for training and 1000 for testing. We set the weight decay to 0.0001 and the momentum to 0.9. We used pretrained ImageNet1k \cite{deng2009imagenet} weights for all backbones. 

\subsection{Comparison with proposed models}
We compare and analyze the proposed models and their variants. In summary, BshapeNet is better than BboxNet, the Scored model is better than the Thick model, and Scored BsahpeNet+ demonstrates the best performance. 

\noindent\textbf{Thickness of Boundaries:}\hspace{.5em}As shown in Tables 2 and 3, the 7-px models demonstrated the best results in both object detection and instance segmentation in COCO. Unlike COCO, the 3-px model, rather than the 7-px, was the best model in Cityscapes (Tables 6 and 7). For COCO, good performance is achieved using relatively thick boundary masks owing to the variety of objects (81 classes), sizes, and backgrounds. That is, it must be thick enough to cover the various scales of the objects. Meanwhile, in Cityscapes, the objects are relatively simpler than those of COCO; a fairly thick boundary affects the model negatively because the boundary dilation introduces noise due to false boundaries.

\noindent\textbf{BshapeNet vs. BboxNet:}\hspace{.5em}The BshapeNet with the same variant condition is significantly more accurate in object detection (Tables 2 and 6). The accuracy of BshapeNet's best model is 42.1 (32.0) $AP_{bb}$, while that of BboxNet's best model is 38.1 (29.7) $AP_{bb}$, as shown in Tables 2 and 6, respectively. This demonstrates that the boundary shape information of the object facilitates in detecting objects much more effectively than the bbox information.

\noindent\textbf{Scored masks vs. Thick masks:}\hspace{.5em}The results of Scored models with the same variant condition surpass those of thick models in MS COCO and Cityscapes, as shown in Tables 2, 3, 6, and 7. In particular, the Scored models are much better than the Thick models in instance segmentation (Tables 3 and 7). For example, 7-px Scored BshapeNet (36.7 $AP_{mk}$) is 5.2 points higher than 7-px Thick BshapeNet (31.5 $AP_{mk}$) in COCO (Table 3). This confirms that filling the false boundary values (distance-based scored values) differently than the true boundary value improves object detection performance.

\noindent\textbf{Scored BshapeNet vs. Scored BshapeNet+:}\hspace{.5em} Because the performance of Scored BshapeNet is the best among our models, Scored BshapeNet+ models with the instance mask branch added to this model were tested and the results are shown in Tables 2$\sim$7, and 9. BshapeNet+ performed better than BshapeNet in all experiments under the same conditions. In our opinion, the instance mask branch has been added to allow for the model to learn more information. The results of this model are analyzed in the next subsection.
\subsection{Object Detection}
\noindent\textbf{Main Results:}\hspace{.5em}All BshapeNet and BshapeNet+ models showed better detection performance than Faster R-CNN$\ddagger$ and Mask R-CNN in both COCO dataset (\texttt{minival} and \texttt{test-dev}) and Cityscapes (\texttt{val}), as shown in Tables 2, 4, 5, and 6. In particular, the best result (42.1 AP) of BshapeNet for COCO was 5.1 points (4.1 points) higher than Faster R-CNN$\ddagger$ (Mask R-CNN) in Table 2. Similarly, in Cityscapes, our best BshapeNet result (32 AP) was 3.1 points (2.4 points) higher than Faster R-CNN$\ddagger$ (Mask R-CNN) (Table 6). These results demonstrate that our mask branches can help improve the performance of object detection. In addition, it shows that the scored bshape mask branch is more effective than the instance mask branch for object detection. BshapeNet+ obtained better results for both data; in more detail, it achieved 42.3 AP (32.3 AP) on COCO \texttt{minival} (on Cityscapes \texttt{val}). 

Table 5 shows that our model achieves very competitive results with state-of-the-art models in MS COCO \texttt{test-dev}. BshapeNet achieves 42.3 AP, and the result of BshapeNet+ is 42.4 AP. BshapeNet+ has the highest performance among the SOTA models presented.

\noindent\textbf{Ablation Studies:}\hspace{.5em}We also performed ablations with COCO \texttt{test-dev} and \texttt{minival} as in Table 4. We compared Faster R-CNN$\ddagger$ with BshapeNet to check the effect of the bshape mask branch. The \texttt{test-dev} result of BshapeNet (42.3 AP) was significantly higher than that (37.3 AP) of Faster R-CNN$\ddagger$. When we changed the backbone to ResNeXt101, BshapeNet showed a score of 42.5 AP in \texttt{test-dev}, which was 0.2 points higher. Adding an instance mask branch to this model improved the detection performance by 0.3 points to 42.8 AP.

\noindent\textbf{Results of Bbox Mask Branch:}\hspace{.5em}The bboxes can be predicted only with our bbox mask of BboxNet, and bbox AP can be calculated using the coordinates of the top left corner and bottom right corner of the predicted box. The results of the bbox mask branch are similar or slightly higher than those of the bbox regressor as in Table 8. In addition, the intersection of the two results improved the accuracy. 

\begin{table}[!t]
	\begin{center}
		
		\resizebox{\linewidth}{!}{
			\begin{tabular}{l|c|cccccccccc}
				\hline\noalign{\smallskip}
				Object Det& backbone&\(AP_{bb}\) &\(AP^{50}_{bb}\) &\(AP^{75}_{bb}\) &\(AP^{L}_{bb}\)  &\(AP^{M}_{bb}\) &\(AP^{S}_{bb}\) \\
				\noalign{\smallskip}
				\hline
				\noalign{\smallskip}
				R-FCN\cite{dai2016r}  &R-101&\large29.9&\large51.9&\large-&\large45.0&\large32.8&\large10.8\\
				\noalign{\smallskip}
				SSD-513\cite{fu2017dssd}  &R-101&\large31.2&\large50.4&\large33.3&\large49.8&\large34.5&\large10.2\\
				\noalign{\smallskip}
			    Faster R-CNN$\ddagger$\cite{he2017mask}&R-101&\large37.3&\large59.6&\large40.3&\large48.8&\large40.2&\large19.8\\
			    \noalign{\smallskip}
			    Mask R-CNN\cite{he2017mask}&R-101&\large38.2&\large60.3&\large41.7&\large50.2&\large41.1&\large20.1\\
			    \noalign{\smallskip}
			    RetinaNet\cite{lin2018focal}&R-101&\large39.1&\large59.1&\large42.3&\large50.2&\large44.2&\large24.1\\
				\noalign{\smallskip}
				\hline
				\noalign{\smallskip}
				BshapeNet  &R-101&\large42.3&\large64.5&\large46.4&\large53.0&\large45.7&\large24.9\\
				\noalign{\smallskip}
				BshapeNet+  &R-101&\large\textbf{42.4}&\large\textbf{64.7}&\large\textbf{46.6}&\large\textbf{53.1}&\large\textbf{45.9}&\large\textbf{24.9}\\
				\noalign{\smallskip}
				\hline
				\hline\noalign{\smallskip}
				Instance Seg & backbone&\(AP_{mk}\) &\(AP^{50}_{mk}\) &\(AP^{75}_{mk}\) &\(AP^{L}_{mk}\)  &\(AP^{M}_{mk}\) &\(AP^{S}_{mk}\) \\
				\noalign{\smallskip}
				\hline
				\noalign{\smallskip}
				MNC\cite{dai2016instance}  &R-101&\large24.6&\large44.3&\large24.8&\large43.6&\large25.9&\large4.7\\
				\noalign{\smallskip}
			    FCIS\cite{shrivastava2016training}&R-101&\large29.2&\large49.5&\large-&\large50.0&\large31.3&\large7.1\\
			    \noalign{\smallskip}
			    FCIS+OHEM\cite{shrivastava2016training}&R-101&\large33.6&\large54.5&\large-&\large-&\large-&\large-\\
			    \noalign{\smallskip}
			     Mask R-CNN\cite{he2017mask}&R-101&\large35.7&\large58.0&\large37.8&\large52.4&\large38.1&\large15.5\\
				\noalign{\smallskip}
				\hline
				\noalign{\smallskip}
				BshapeNet  &R-101&\large37.0&\large58.1&\large38.2&\large51.9&\large38.5&\large16.7\\
				\noalign{\smallskip}
				BshapeNet+  &R-101&\large\textbf{37.5}&\large\textbf{59.2}&\large\textbf{39.5}&\large\textbf{53.1}&\large\textbf{39.9}&\large\textbf{17.3}\\
				\noalign{\smallskip}
				\hline
		\end{tabular}}
	\end{center}
	\caption{Comparison of object detection and instance segmentation results of 7-px Scored BshapeNet with the state of the art models on COCO \texttt{test-dev}.}
\end{table}
\begin{table}[t]
\begin{center}

\resizebox{\linewidth}{!}{
\begin{tabular}{c|l|ccc|l|ccc}
\hline\noalign{\smallskip}
\multicolumn{2}{l|}{Model (R-50)}&  \(AP_{bb}\) &\(AP^{50}_{bb}\) &\(AP^{75}_{bb}\) & Model (R-50) & \(AP_{bb}\) &\(AP^{50}_{bb}\) &\(AP^{75}_{bb}\) \\
\noalign{\smallskip}
\hline
\noalign{\smallskip}
\multirow{5}{*}{{\rotatebox[origin=c]{90}{\large{T h i c k}}}}
&BboxNet (3) &\Large29.3&\Large	48.4&\Large28.6&BshapeNet (3)  &\Large30.3&\Large49.7&\Large28.7\\
\noalign{\smallskip}
&BboxNet (5)  &\Large29.0&\Large48.0&\Large28.5&BshapeNet (5) &\Large30.0&\Large49.4&\Large28.6\\
\noalign{\smallskip}
&BboxNet (7) &\Large29.2&\Large48.1&\Large28.3&BshapeNet (7) &\Large29.9&\Large49.2&\Large28.6\\
\noalign{\smallskip}
&BboxNet (11)  &\Large29.1&\Large48.2&\Large28.1&BshapeNet (11)&\Large29.9&\Large49.2&\Large28.7\\
\noalign{\smallskip}
\hline
\noalign{\smallskip}
\multirow{7}{*}{{\rotatebox[origin=c]{90}{\large{S c o r e d}}}}
&BboxNet (3) &\Large29.7&\Large48.9&\Large28.9&\textbf{BshapeNet (3)} &\Large\textbf{32.0}&\Large\textbf{52.0}&\Large\textbf{32.1}\\
\noalign{\smallskip}
&BboxNet (5) &\Large29.2&\Large48.1&\Large28.8&BshapeNet (5) &\Large31.4&\Large50.5&\Large29.9\\
\noalign{\smallskip}
&BboxNet (7) &\Large29.4&\Large48.2&\Large28.3&BshapeNet (7)&\Large30.9&\Large50.1&\Large29.2\\
\noalign{\smallskip}
&BboxNet (11) &\Large29.4&\Large48.2&\Large28.5&BshapeNet (11)&\Large30.7&\Large50.3&\Large29.1\\
\noalign{\smallskip}
\cline{2-9}
\noalign{\smallskip}
&\textbf{BshapeNet+ (3)} &\Large\textbf{32.3}&\Large\textbf{52.4}&\Large\textbf{32.5}&BshapeNet+ (7)&\Large31.3&\Large50.8&\Large29.7\\
&BshapeNet+ (5) &\Large31.8&\Large51.7&\Large31.6&BshapeNet+ (11)&\Large31.2&\Large51.0&\Large29.7\\
\noalign{\smallskip}
\hline
\noalign{\smallskip}
\multicolumn{2}{c|}{Faster R-CNN$\ddagger$ \small{(ours)}}&\Large28.9&\Large48.4 &\Large28.1&Mask  R-CNN \small{(ours)}&\Large29.6&\Large49.1 &\Large29.2\\
\noalign{\smallskip}
\hline
\end{tabular}}
\end{center}
\caption{Object detection results on Cityscapes \texttt{val} dataset with bbox AP (\%).}

\end{table}
\begin{table}[!t]
\begin{center}
\resizebox{\linewidth}{!}{
\begin{tabular}{c|l|ccc|l|ccc}
            \hline\noalign{\smallskip}
                \multicolumn{2}{l|}{Model (R-50)}&  \(AP_{mk}\) &\(AP^{50}_{mk}\) &\(AP^{75}_{mk}\)&Model (R-50)&  \(AP_{mk}\) &\(AP^{50}_{mk}\) &\(AP^{75}_{mk}\)\\
				\noalign{\smallskip}
				\hline
				\noalign{\smallskip}\noalign{\smallskip}
				\multirow{2}{*}{{\rotatebox[origin=c]{90}{T h i c k}}}
				&BshapeNet (3)  &\Large{29.4}&\Large{48.2}&\Large{29.0}&BshapeNet (7)&\Large{29.1}&\Large{47.5}&\Large{29.0}\\
				\noalign{\smallskip}\noalign{\smallskip}
				&BshapeNet (5) &\Large{29.4}&\Large{48.0}&\Large{29.1}&BshapeNet (11)&\Large{28.9}&\Large{47.6}&\Large{28.8}\\
  				\noalign{\smallskip}\noalign{\smallskip}
				\hline
				\noalign{\smallskip}
				\multirow{5}{*}{{\rotatebox[origin=c]{90}{S c o r e d}}}
				&\textbf{BshapeNet (3)} &\Large{\textbf{32.1}}&\Large{\textbf{49.8}}&\Large{\textbf{30.2}}&BshapeNet (7)&\Large{31.7}&\Large{49.0}&\Large{29.6}\\
				\noalign{\smallskip}\noalign{\smallskip}
				&BshapeNet (5) &\Large{31.9}&\Large{49.5}&\Large{29.9}&BshapeNet (11)&\Large{31.4}&\Large{48.7}&\Large{29.6}\\
				\noalign{\smallskip}
				
				\cline{2-9}
				\noalign{\smallskip}
				&\textbf{BshapeNet+ (3)} &\Large{\textbf{33.5}}&\Large{\textbf{50.7}}&\Large{\textbf{30.7}}&BshapeNet+ (7)&\Large{32.0}&\Large{50.2}&\Large{29.7}\\
				\noalign{\smallskip}\noalign{\smallskip}
				&BshapeNet+ (5)  &\Large{32.3}&\Large{50.7}&\Large{30.2}&BshapeNet+ (11)&\Large{31.9}&\Large{48.8}&\Large{29.5}\\
				\noalign{\smallskip}
				\hline
				\noalign{\smallskip}
				\multicolumn{2}{l|}{Mask R-CNN \cite{he2017mask}}&\Large{31.5}&\Large{-}&\Large{-}&Mask R-CNN (ours)&\Large{31.2}&\Large{49.7} &\Large{29.6}\\
				\noalign{\smallskip}
				
				\hline
		\end{tabular}}
	\end{center}
	\caption{Instance segmentation results on Cityscapes \texttt{val} dataset with mask AP (\%).}
\end{table}

\begin{table}[t]
	\begin{center}
		\resizebox{8cm}{!}{
			\begin{tabular}{l|cccccc|ccc}
				\hline\noalign{\smallskip}
				\multirow{2}{*}{Model} &\multicolumn{6}{|c}{COCO (\texttt{minival})}&\multicolumn{3}{|c}{Cityscapes (\texttt{val})}\\
				& \(AP_{bb}\) &\(AP^{50}_{bb}\) &\(AP^{75}_{bb}\)  &\(AP^{L}_{bb}\)&\(AP^{M}_{bb}\)&\(AP^{S}_{bb}\)&\(AP_{bb}\) &\(AP^{50}_{bb}\) &\(AP^{75}_{bb}\) \\
                \noalign{\smallskip}
                \hline
                \noalign{\smallskip}
                 BBR&\Large38.1&\Large59.7&\Large40.9&\Large50.0&\Large41.1&\Large20.3&\Large29.7&\Large48.9&\Large28.9\\
                \noalign{\smallskip}
                 BM&\Large38.2&\Large59.9&\Large40.8&\Large49.7&\Large42.3&\Large20.9&\Large29.7&\Large49.1&\Large28.8\\
                \noalign{\smallskip}
                 BBR $\cap$ BM &\Large38.4&\Large60.1&\Large42.4&\Large50.3&\Large42.5&\Large21.2&\Large29.9&\Large49.2&\Large28.8\\
                \noalign{\smallskip}
				\hline
				
		\end{tabular}}
	\end{center}
	\caption{Object detection results of Scored BboxNet (ResNet101) on COCO \texttt{minival} and object detection results of Scored BboxNet (ResNet50) on Cityscapes \texttt{val}. BBR is the result from the bbox regressor. BM is the result from the bbox mask. BBR $\cap$ BM is the result from the intersection of BM and BBR.}
\end{table}
\subsection{Instance Segmentation}

\noindent\textbf{Main Results:}\hspace{.5em}We present the results of instance segmentation of BshapeNet+ and BshapeNet using \texttt{minival} and \texttt{test-dev} in COCO and \texttt{val} and \texttt{test-server} in Cityscapes in Tables 3, 4, 5, 7, and 9. In Tables 3 and 7, we show that all BshapeNet+ models (except one model) obtain superior performance over Mask R-CNN in both datasets. In particular, BshapeNet+, which showed the best performance, achieved 37.1 AP and 33.5 AP in COCO and Cityscapes, respectively. Compared to the Mask R-CNN paper results, we obtained, with our BshapeNet+, an improvement of 1.7 points in COCO and an improvement of 2.0 points in Cityscapes and the result of BshapeNet is 1.3 points higher in COCO and 0.6 points higher. 

Our models achieved very comparable results with the SOTA models in COCO \texttt{test-dev} and Cityscapes \texttt{test-server} as shown in Tables 5 and 9. BshapeNet+ shows a higher instance segmentation performance than the other state-of-the-art models in COCO and Cityscapes. In addition, BshapeNet+ has good performance in small objects as shown in Table 5. BshapeNet+ has a performance of 17.3 AP for small objects and is superior to the current SOTA models.

\noindent\textbf{Ablation Studies:}\hspace{.5em}In Table 4, we compare BshapeNet and BshapeNet+ to Mask R-CNN and examine the instance segmentation effect of the bshape mask branch on MS COCO \texttt{test-dev} and \texttt{minival} datasets. When using ResNet101 as a backbone, BshapeNet shows 37.0 AP, which is higher than the Mask R-CNN paper result by 1.3 points in \texttt{test-dev}. With ResNeXt101 as a backbone, BshapeNet has a score of 37.2 AP in \texttt{test-dev}, and it is 0.3 points better than Mask R-CNN (ours). However, adding the instance mask branch to BshapeNet shows a score of 37.9 AP in \texttt{test-dev}, and it is 1.0 points better than Mask R-CNN (ours). 
\begin{table*}[t]
\begin{center}
\resizebox{12cm}{!}{
			\begin{tabular}{l|c|cc|cccccccc}
				\hline\noalign{\smallskip}
				Instance Seg &train dataset&\(AP_{mk}\) &\(AP^{50}_{mk}\) &Person&Rider&Car&Truck&Bus&Train&mcycle&bicycle \\
				\noalign{\smallskip}
				\hline
				\noalign{\smallskip}
				InstanceCut\cite{kirillov2017instancecut} &\texttt{fine+coarse} &\large13.0&\large27.9&\large10.0&\large8.0&\large23.7&\large14.0&\large19.5&\large15.2&\large9.3&\large	4.7\\
				\noalign{\smallskip}
			    DWT\cite{bai2017deep}&\texttt{fine}&\large15.6&\large30.0&\large15.1&\large11.7&\large32.9&\large17.1&\large20.4&\large15.0&\large7.9&\large4.9\\
			    \noalign{\smallskip}
			    BAIS\cite{hayder2017boundary}&\texttt{fine}&\large17.4&\large36.7&\large14.6&\large12.9&\large35.7&\large16.0&\large23.2&\large19.0&\large10.3&\large7.8\\
			    \noalign{\smallskip}
			    DIN\cite{arnab2017pixelwise}&\texttt{fine+coarse}&\large20.0&\large38.8&\large16.5&\large16.7&\large25.7&\large20.6&\large30.0&\large23.4&\large17.1&\large10.1\\
			    \noalign{\smallskip}
			    SGN\cite{liu2017sgn}&\texttt{fine+coarse}&\large25.0&\large44.9&\large21.8&\large20.1&\large39.4&\large24.8&\large33.2&\large\textbf{30.8}&\large17.7&\large12.4\\
			    \noalign{\smallskip}
			    PolygonRNN++\cite{acuna2018efficient}&\texttt{fine}&\large25.4&\large45.5&\large29.4&\large21.8&\large48.3&\large21.1&\large32.3&\large23.7&\large13.6&\large13.7\\
			    \noalign{\smallskip}
			    Mask R-CNN\cite{he2017mask}&\texttt{fine}&\large26.2&\large49.9&\large30.5&\large\textbf{23.7}&\large46.9&\large22.8&\large32.2&\large18.6&\large19.1&\large\textbf{16.0}\\
			    \noalign{\smallskip}
				\hline
				\noalign{\smallskip}
				BshapeNet  &\texttt{fine} &\large27.1&\large50.3&\large29.6&\large23.3&\large46.8&\large25.8&\large32.9&\large24.6&\large20.3&\large14.0\\
				\noalign{\smallskip}
				BshapeNet+&\texttt{fine} &\large\textbf{27.3}&\large\textbf{50.5}&\large\textbf{30.7}&\large23.4&\large\textbf{47.2}&\large\textbf{26.1}&\large\textbf{33.3}&\large24.8&\large\textbf{21.5}&\large14.1\\
				\noalign{\smallskip} 
				\hline 
\end{tabular}}
	\end{center}
	\caption{Comparison of instance segmentation results of BshapeNet  with the state-of-the art models on Cityscapes \texttt{test-server}. The 3-px Scored BshapeNet was used.}
\end{table*}
\begin{figure*}[t]
  \centering    
  \includegraphics[width=140mm]{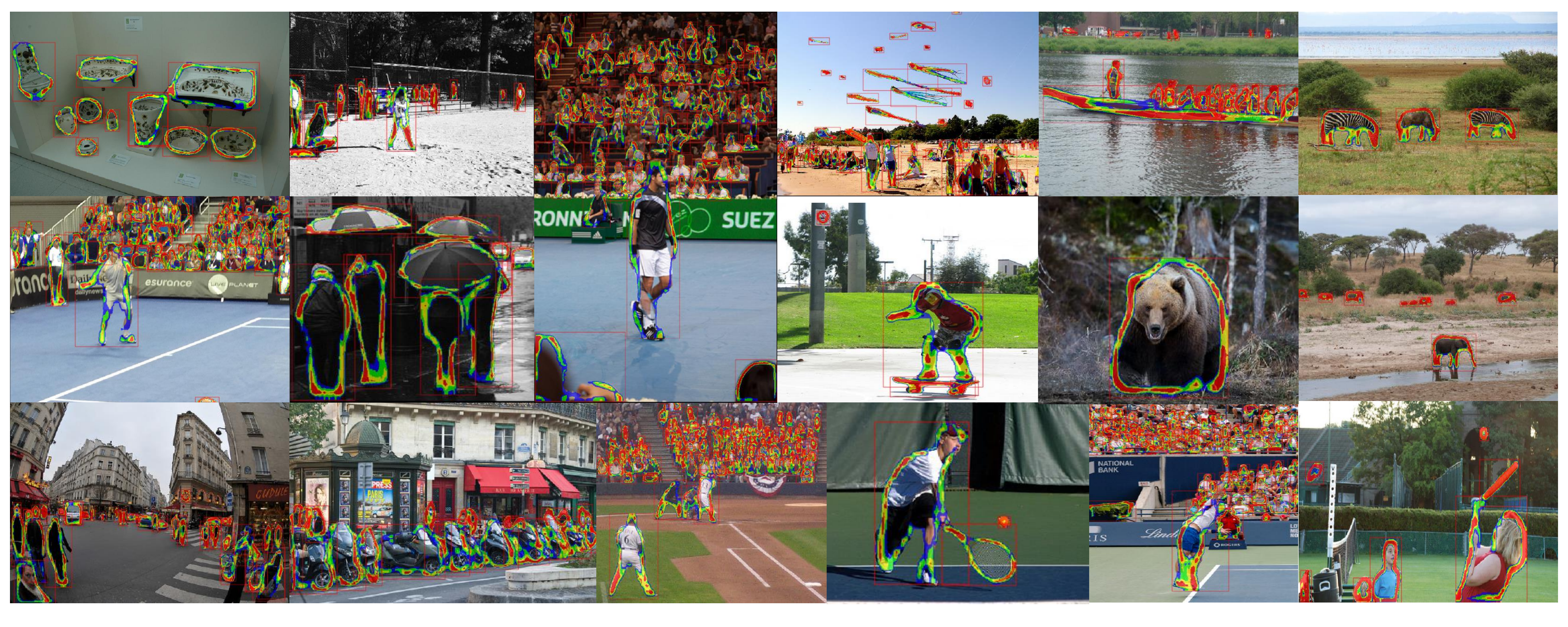}
  \caption{Examples of object detection and instance segmentation results for the 7-px
  Scored BshapeNet (ResNet101) on the MS COCO val dataset. Scored BshapeNet detects the outlines of
  objects and the bounding boxes well.}
\end{figure*}
\begin{figure*}[!t]
	\centering    
	\includegraphics[width=160mm]{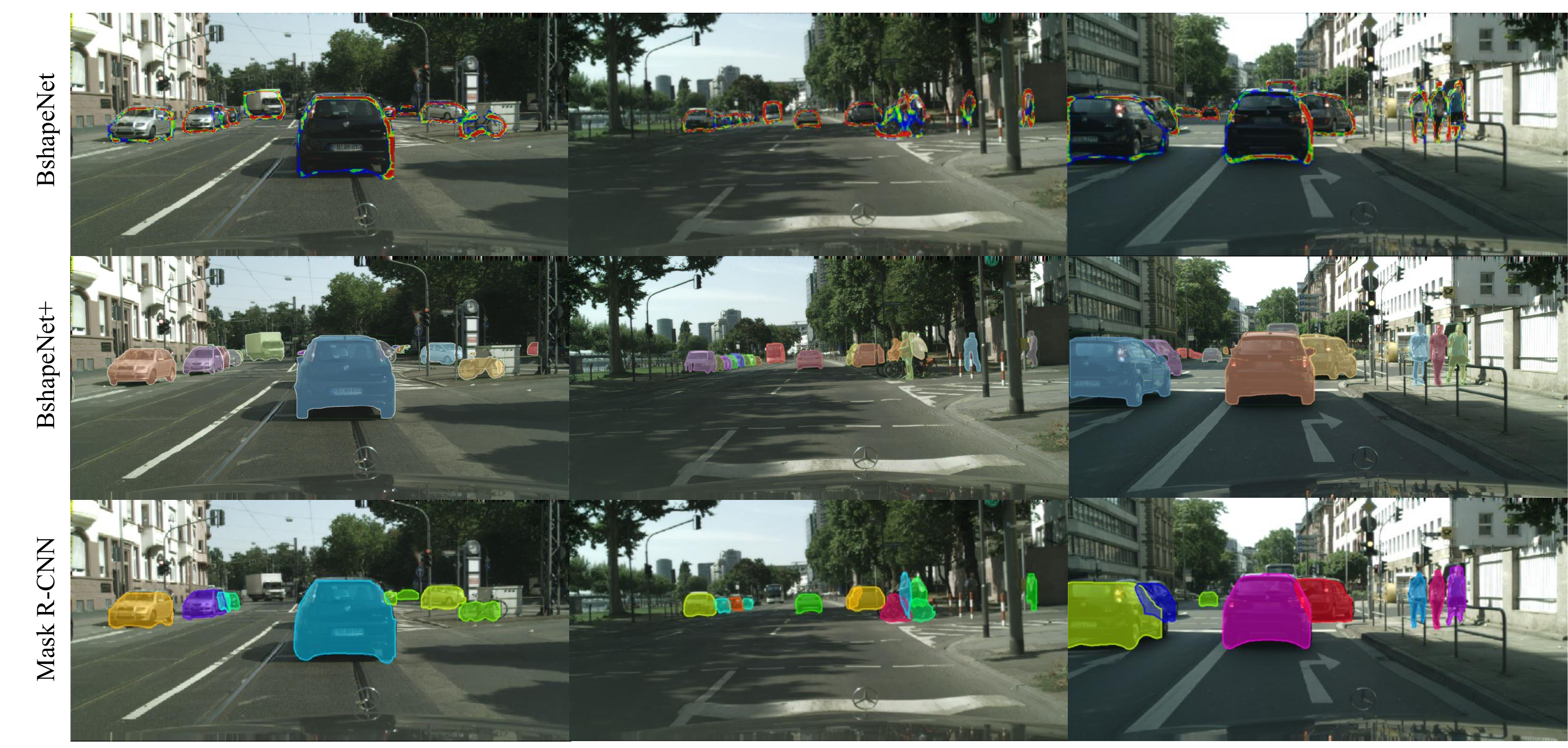}
	\caption{Comparison of the results of instance segmentation of Cityscapes images by the proposed model (3-px Scored BshapeNet (top) and 3-px Scored BshapeNet+ (middle)), and Mask R-CNN (bottom). Both models are trained in the same experimental environment and under the ResNet50 backbone. The proposed models detect objects that are not detected by Mask R-CNN as well as occluded objects.}
\end{figure*}

\section{Conclusion}
We demonstrated a significantly improved performance by additionally providing the locations of objects with a different format as well as the coordinates to object detection algorithms. Further, our methods could be applied easily to all detection algorithms using bbox regression. Our branch was removable at inference without performance degradation; therefore, it is an effective method to reduce computational cost. The experimental results of BshapeNet+ demonstrated that the proposed method improved the performance of instance segmentation, and that it was particularly good for small objects.

\section{Acknowledgments}
This research was supported by System LSI Business, Samsung Electronics Co., Ltd.

{\small
\bibliographystyle{ieee}
\bibliography{BshapeNet}
}

\end{document}